\documentclass[10pt,twocolumn,letterpaper]{article}
\usepackage{nopageno}
\usepackage{3dv}
\usepackage{times}
\usepackage{epsfig}
\usepackage{graphicx}
\usepackage{amsmath}
\usepackage{amssymb}
\usepackage{multirow}
\usepackage{xcolor}
\definecolor{green(ryb)}{rgb}{0.4, 0.69, 0.2}
\usepackage[export]{adjustbox}
\renewcommand{\paragraph}{\textbf}
\usepackage[pagebackref=true,breaklinks=true,letterpaper=true,colorlinks,bookmarks=false]{hyperref}

\newcommand{\VEC}{\operatorname{vec}}
\newcommand{\zerovec}{\mathbf{0}}
\newcommand{\matI}{\mathbf{I}}
\newcommand{\R}{\mathbb{R}}
\newcommand{\N}{\mathbb{N}}
\newcommand{\SO}{\text{SO}}
\newcommand{\local}{\text{rel}}
\newcommand{\canonical}{\text{can}}

\newcommand{\trace}{\operatorname{tr}}

\makeatletter
\@tfor\next:=abcdefghijklmnopqrstuvwxyz\do{%
  \def\command@factory#1{%
    \expandafter\def\csname vec#1\endcsname{\mathbf{#1}}
  }
 \expandafter\command@factory\next
}

\@tfor\next:=ABCDEFGHIJKLMNOPQRSTUVWXYZ\do{%
  \def\command@factory#1{%
    \expandafter\def\csname mat#1\endcsname{\mathbf{#1}}
  }
 \expandafter\command@factory\next
}

\@tfor\next:=ABCDEFGHIJKLMNOPQRSTUVWXYZ\do{%
  \def\command@factory#1{%
    \expandafter\def\csname set#1\endcsname{\mathcal{#1}}
  }
 \expandafter\command@factory\next
}

\def\greekvectors#1{%
 \@for\next:=#1\do{%
    \def\X##1;{%
     \expandafter\def\csname mat##1\endcsname{\boldsymbol{\csname##1\endcsname}}
     }
   \expandafter\X\next;
  }
}
\greekvectors{alpha,beta,iota,gamma,lambda,nu,eta,Gamma,varsigma,Phi,Theta}

\makeatother

\threedvfinalcopy %

\def\threedvPaperID{0056} %

\ifthreedvfinal\fi
\begin{document}
\title{Convex Optimisation for Inverse Kinematics}

\author{Tarun Yenamandra\textsuperscript{1,2}, Florian Bernard\textsuperscript{1,2}, Jiayi Wang\textsuperscript{1,2}, Franziska Mueller\textsuperscript{1,2}, and Christian Theobalt\textsuperscript{1,2} \\
\textsuperscript{1}MPI Informatics, \textsuperscript{2}Saarland Informatics Campus
}

\maketitle
\begin{abstract}
  We consider the problem of inverse kinematics (IK), where one wants to find the parameters of a given kinematic skeleton that best explain a set of observed 3D joint locations. The kinematic skeleton has a tree structure, where each node is a joint that has an associated geometric transformation that is propagated to all its child nodes. The IK problem has various applications in vision and graphics, for example for tracking or reconstructing articulated objects, such as human hands or bodies. Most commonly, the IK problem is tackled using local optimisation methods. A major downside of these approaches is that, due to the non-convex nature of the problem, such methods are prone to converge to unwanted local optima and therefore require a good initialisation. In this paper we propose a convex optimisation approach for the IK problem based on semidefinite programming, which admits a polynomial-time algorithm that globally solves (a relaxation of) the IK problem. Experimentally, we demonstrate that the proposed method significantly outperforms local optimisation methods using different real-world skeletons.
\end{abstract}
\section{Introduction}

The inverse kinematics (IK) problem plays an important role in robotics, computer games, graphics, and vision, as it is a fundamental building block for animating, controlling, tracking and reconstructing articulated objects, such as robotic arms or human bodies.
The IK problem refers to the task of recovering parameters of a kinematic skeleton (e.g.~joint angles), given a set of observed locations for (some of) the joints.
\begin{figure}
    \includegraphics[width=\linewidth]{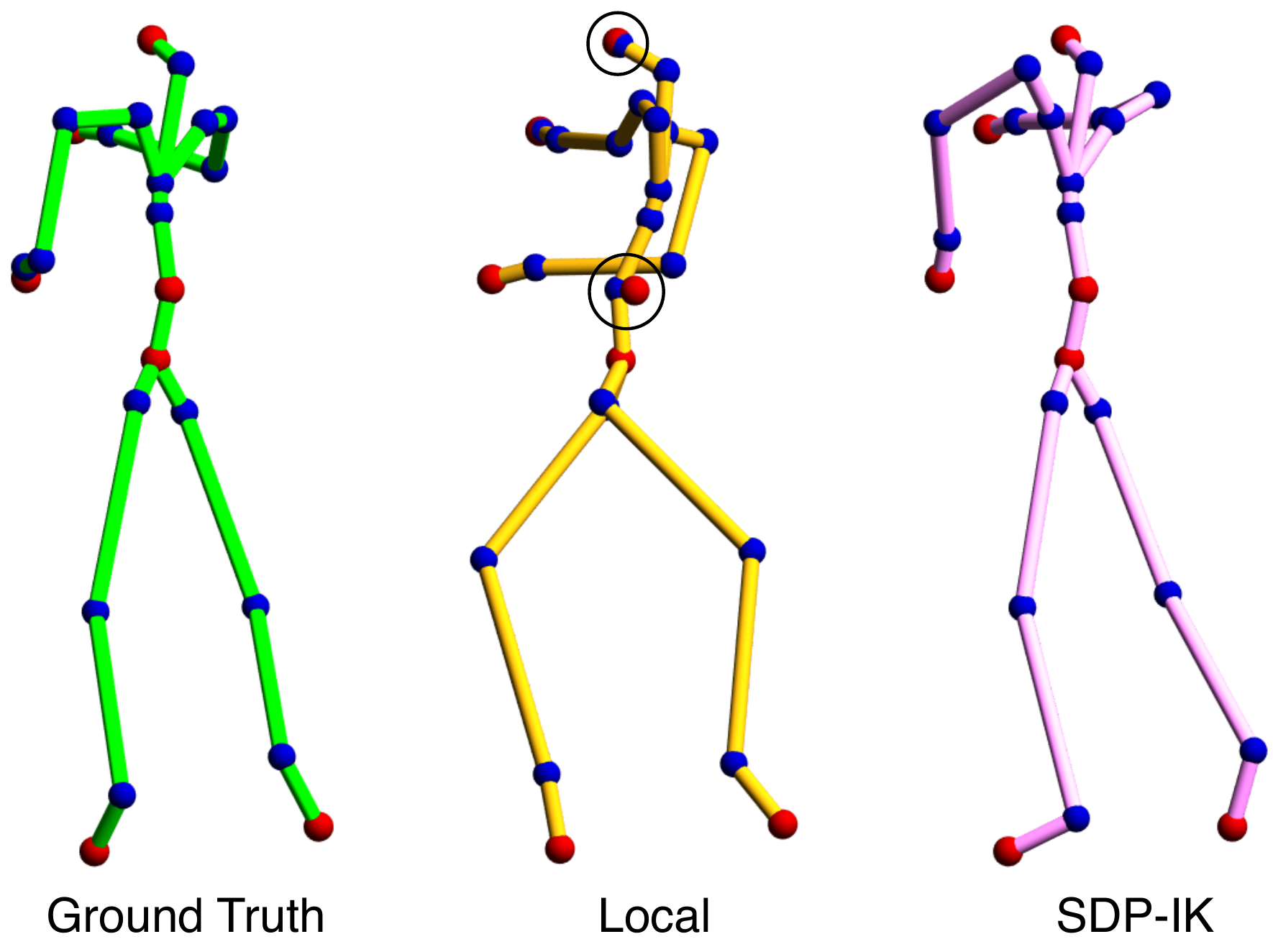}
    \caption{Our \texttt{SDP-IK} method results in a better fit compared to local optimisation.  Left: Ground truth pose, where the observed points are shown in red. Centre: Local IK methods  can get stuck in an unwanted local minimum. Right: Our convex \texttt{SDP-IK} formulation can reliably reach a good optimum that is much closer to the ground truth.}
    \label{fig:teaser} 
\end{figure}
The kinematic skeleton is represented as a tree, where each node is a joint that has translational and/or rotational degrees-of-freedom. In forward kinematics, one computes the positions of the joints given the translation and rotation parameters. Here, the transformation for each joint is defined relative to its parent joint, so that a chain of transformations is applied along the path from the root node to a leaf node (end-effector). Inverting this process,~i.e.~given some of the joint positions, one wants to recover the kinematic parameters that lead to this joint position configuration.

The IK problem is considered to be difficult due to several reasons. On the one hand, depending on the kinematic skeleton and the observed joint positions, the inverse kinematics can be ill-posed so that there may be multiple configurations of kinematic parameters that lead to the same joint positions.
On the other hand, depending on the given joint locations, a solution that produces an exact fit to the observations might not exist.
Lastly, the resulting optimisation problem is non-convex, which makes it generally difficult to find a globally optimal solution. Hence, approximations are oftentimes used in practice, which in turn require an initialisation that is sufficiently close to a global optimum. In computer vision, one of the most dominant applications of the IK problem is tracking and reconstructing articulated objects based on a temporal sequence of data (e.g.~depth images, or RGB images). A common approach for tackling tracking applications is to initialise the kinematic para\-meters for the next frame using the tracked result from the current frame, and then solve the IK problem using local optimisation methods. However, this is not possible for the very first frame of a sequence. Hence, in many works the authors assume that a good initialisation is available for the first frame,~e.g.~based on an initial calibration with a neutral pose, as done in~\cite{ellipsoidtracker_3dv2014}. In contrast, our proposed method is entirely initialisation-free, and is therefore well-suited for handling such cases. The main contribution of this work is a polynomial-time solution of the IK problem, coined \texttt{SDP-IK}, that finds a global optimum of a (convex relaxation of) the IK problem based on semidefinite programming.
\section{Related Work}
In this section we first address local optimisation methods for the inverse kinematics problem, followed by global methods. Subsequently, we summarise the most relevant works that consider semidefinite programming relaxations of related optimisation problems.

\paragraph{Local IK approaches:}
Local optimisation methods seek to iteratively find a solution of the IK problem based on a given initial estimate. One class of such methods use the first-order Taylor approximation of the problem and attempt to solve a linear system  characterised by the Jacobian matrix.
Alternatives are the Moore-Penrose pseudoinverse method, the Jacobian transpose method (equivalent to gradient descent for least-squares error), the Levenberg-Marquardt method (equivalent to gradient descent for damped least-squares error), and other variants~\cite{Abderrahim2013, BussIK, Torras2012,  Kenwright2012}. Second-order methods also exist, but such approaches require the computation of the Hessian of the forward kinematics function, which incur higher computational cost. Quasi-Netwon methods, such as the Broyden-Fletcher-Goldfarb-Shanno (BFGS) algorithm, have been used to provide faster approximations for solving the IK problem~\cite{Chin97}. 

Instead of trying to approximate and invert the forward kinematics function, heuristic methods employ simple rules to be iteratively followed and can often reach the IK solution. Cyclic coordinate descent (CCD) and its variants are heuristic methods that seek to minimise joint errors by changing one kinematic parameter at a time~\cite{Merrick:2004, Wang1991}. Forward and backward reaching IK (FABRIK)~\cite{Aristidou:2011:FABRIK} provides a method that can provably converge to the correct solution, when feasible, for a single unconstrained kinematic chain \cite{Aristidou2016}. Evolutionary algorithms such as particle swarm optimisation and genetic algorithms, are heuristics inspired by evolution and are used to solve the IK problem in~\cite{starke2016efficient,starke2017memetic}.

Due to their speed and ease-of-use, local optimisation methods are widely used in applications such as character animation \cite{Kenwright2012}, motion re-targeting \cite{Hecker:2008:RMR:1399504.1360626}, model-based tracking \cite{ellipsoidtracker_3dv2014, Tkach:2016}, post-processing on pose estimation \cite{VNect_SIGGRAPH2017, GANeratedHands_CVPR2018}, and data visualisation \cite{Merrick:2004}. However, despite their popularity, local IK approaches require a good initialisation for complex, real-world skeletons, as otherwise they are prone to converge to unwanted local optima.

\paragraph{Global IK methods:}
Global IK methods aim to avoid this tendency getting stuck in local optima, and instead seek to obtain a global solution.
One way of of achieving initialisation independence is based on training a machine learning model to solve the IK problem. In~\cite{Bocsi:STRUCTIK:2011}, the authors learn an inverse kinematics function which maps from joint locations to kinematic parameters using a structured learning method. 

An alternative to learning approaches are global optimisation methods.
In~\cite{dai2018synthesis}, the authors tackle the problem of force-closure grasp synthesis based on sequential semidefinite programming, where rotation constraints are modelled in terms of bilinear matrix inequalities involving quaternions and rotation matrices.  An approach for a feasibility formulation of the IK problem based on mixed-integer programming (MIP) is presented in~\cite{dai2017global}. 
The main idea here is to discretise all non-convex constraints based on binary variables. The resulting problem is then solved with a branch and bound algorithm, which is known to have exponential worst-case time complexity. 

Contrary to the discussed works, which address feasibility versions of the IK problem, and/or do not admit polynomial-time algorithms, we propose a principled polynomial-time approach for the IK problem. Moreover, we consider a least-squares version of the IK problem, which is most relevant for the majority of applications in vision and graphics, such as for the tracking or the reconstruction of articulated objects.

\paragraph{Semidefinite programming relaxations:}
Semidefinite programming (SDP) relaxations are a popular way for tackling non-convex optimisation problems. Such methods have been successfully used for a range of different problems in vision and beyond,~e.g.~for graph matching~\cite{Schellewald:2005up, Zhao:1998wc} or multi-graph matching~\cite{bernard:2018,kezurer2015}, the rigid registration of point-clouds~\cite{Khoo:2016hr,Maron:2016vv}, the segmentation of images~\cite{Wang:2013vq}, or for permutation synchronisation~\cite{ChenGuibasHuang14_NearOptimalJointObjectMatchingViaConvexRelaxation}. However, generally such approaches are computationally expensive, since many of the SDP relaxations are based on a \emph{lifting} of the variables, so that the size of the optimisation problem increases quadratically when moving from the original non-convex problem to the convex relaxation~\cite{kezurer2015,Schellewald:2005up,Zhao:1998wc}.

One scenario where SDP relaxations particularly stand out is in problems involving 3D rotation matrices. On the one hand, lifting a matrix variable of size $3{\times}3$ merely results in a relatively small variable of size $1{+}(3{\cdot}3)^2{=}82$, so that such problems can be solved efficiently.
On the other hand, some relaxations that involve a single rotation matrix have been observed to be tight in practice,~i.e.~even when solving a relaxation of a non-convex problem, the so-found solution is a global minimiser of the original non-convex problem. This has for example been empirically demonstrated in~\cite{Briales:un} for the registration of 3D objects (with known correspondence). Other approaches that consider SDP approaches for problems involving rotations have been demonstrated for SLAM~\cite{Rosen:2015dv}, pose-graph optimisation~\cite{Carlone:2018ji}, or rotation averaging~\cite{Bandeira:2014wy,Eriksson:2018cq,Wang:2013tq}.

In our work we consider an SDP relaxation for the inverse kinematics problem, which involves a composition of several rotations that are propagated through the kinematic chain.

\subsection{Notation}
Here, we briefly outline the used notation.  By $\matI_d$ we denote the $d{\times}d$ identity matrix, by $\zerovec_d \in \R^d$ we denote the vector of all zeros, and the operator $\VEC(X)$ vectorises a given input matrix $X$ by concatenating all the columns of $X$.
For an integer $n$ we use the notation $[n]:=\{1,\ldots,n\}$. For a matrix $X$ we write $X_{:,i}$ to denote the vector that is formed by the $i$-th column of $X$, and analogously $X_{i,:}$ to denote the row vector that is formed by the $i$-th row of $X$.
Moreover, for a 3D vector $x \in \R^3$, or a  matrix $T \in \R^{3{\times}3}$, we use $\hat{x} = [x^T~1]^T \in \R^4$ and $\hat{T}\in \R^{4{\times}4}$ to denote their respective representation in homogeneous coordinates. For $X$ being a matrix, the notation $X {\succeq} 0$ means that $X$ is symmetric positive semidefinite.
\section{Inverse Kinematics}
In this section we describe our approach for tackling the inverse kinematics problem. To this end, we first define the forward kinematics model, which is followed by the precise statement of the problem. 
\begin{figure}
    \includegraphics[width=\linewidth]{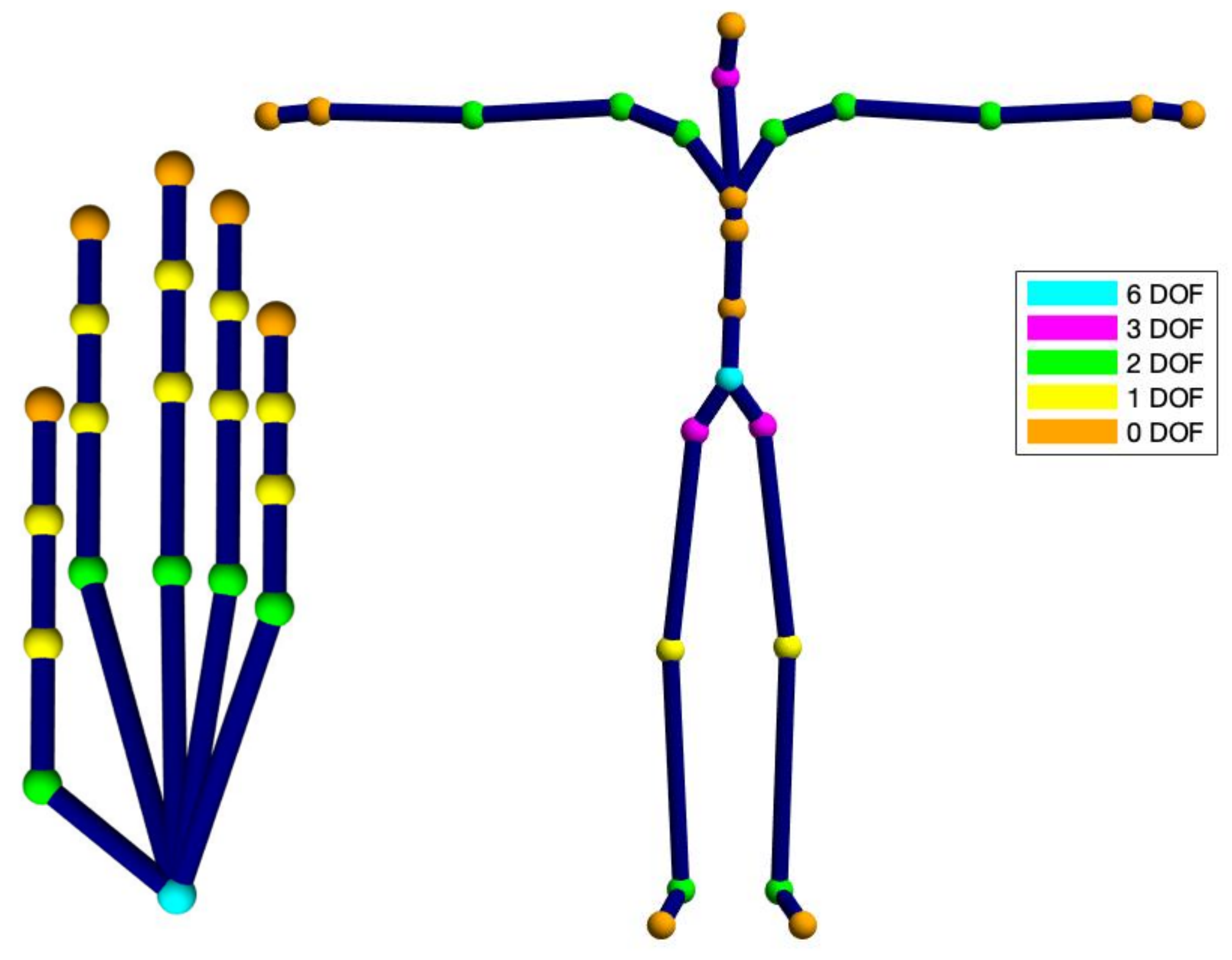}
    \caption{Illustration of a kinematic skeleton of a human hand (left) and a human body (right). The root joint for both skeletons admits 6 DOF (global translation and rotation), whereas all other joints admit between 0 and 3 rotational DOF.}
    \label{fig:kinskel} 
\end{figure}

\subsection{Forward Kinematics Model}\label{sec:fwkin}
We assume that we are given a tree that defines the kinematic skeleton, see Fig.~\ref{fig:kinskel}. The root of the tree has 6 degrees-of-freedom (DOF), of which 3 account for the global rotation, and 3 account for the global translation. Each non-root joint has between 0 and 3 rotational DOF, where each (non-zero) DOF accounts for a rotation of a given angle around a given axis. Joints without DOF are used to model the end-effectors,~e.g.~the fingertips in the human hand,~cf.~Fig.~\ref{fig:kinskel}. 

\paragraph{One-DOF representation:}
W.l.o.g., for convenience, we redefine this generic kinematic skeleton in such a way that each (non-end-effector) joint only has a single DOF that allows for a rotation around a given axis.
To this end, for each joint of the original skeleton that has more than 1 DOF, we simply introduce one or two additional auxiliary joints (placed at the same position) that account for the additional DOF. We emphasise that while this redefinition of the kinematic skeleton does not change its kinematic behaviour, such a representation is more convenient for defining our SDP relaxation, as we will become apparent in Sec.~\ref{sec:sdpik}.

\paragraph{Kinematic skeleton:} Let $J \in \N$ be the total number of joints, where each joint now has at most 1 DOF (due to the redefinition of the skeleton). 
The global translation of the root is denoted by $t \in \R^3$. For each subsequent joint transformation,
let $a_j \in \R^3$ for $j \in [J]$ denote the %
(unit-length) rotation axis of the $j$-th joint, and let $v_j \in \R^3$ be the ``bone-vector'' of the $j$-th joint,~i.e.~the offset of joint $j$ in the coordinate system of its parent (cf.~Fig.~\ref{fig:kinskel}).
For $\theta_j \in \R$ being the parameter of the $j$-th joint, by $\hat{T}(a_j,\theta_j) \in \R^{4 \times 4}$ we denote the transformation from the coordinate system of joint $j$ to the coordinate system of its parent, represented in homogeneous coordinates. 
To be more specific, we have
\begin{align}
    \hat{T}(a_j,\theta_j) = \begin{bmatrix} K(a_j,\theta_j) & v_j \\
                    \zerovec_3^T & 1 \end{bmatrix}\,,
\end{align}
where the rotation matrix $K(a,\theta)$ is obtained by Rodrigues' rotation formula as 
\begin{align}
    K(a,\theta) = \matI_3 {+} \sin(\theta) [a]_{\times} {+} (1{-}\cos(\theta))[a]_{\times}^2 \,.
\end{align}

Here, $[\cdot]_{\times}$ is the skew-symmetric operator that generates a $3{\times}3$ matrix from a 3D vector (i.e.~for $x,y \in \R^3$ we have $[x]_{\times} y = x \times y$). For joints with $0$ DOF (i.e.~for end-effectors), we define $K(a,\theta)$ to be the identity matrix $\matI_3$.

\paragraph{Forward model:}
For $\matTheta := [t^T, \theta_1, \ldots, \theta_J]^T \in \R^{3{+}J}$ being the  parameter vector that stacks all joint parameters, the forward model for computing the position $x_j(\matTheta)$ of the $j$-th joint is given by
\begin{align}
    \hat{x}_j(\matTheta) = \hat{t} + \left[ \prod_{i \in \alpha_j^1}  \hat{T}(a_i,\theta_i) \right] \hat{\zerovec}_3 \,,
\end{align}
where by $\alpha_m^n$ we denote the path from the $m$-th joint to the $n$-th joint (from children to root) in the kinematic skeleton, and $\hat{\zerovec}_3 = [0~0~0~1]^T$ is the zero vector represented in homogeneous coordinates.
For brevity, we will refer to all elements of $\matTheta$ as \emph{angles}, even if they represent translations.

\paragraph{Joint angle constraints:} In addition, for each joint there is an interval $\setI_j$ that defines the range of valid values for $\theta_j$ so that it must hold that $\theta_j \in \setI_j$. For notational convenience, we define $\setI := \R^3 \times \setI_1 \times \ldots \times \setI_J$ and write $\matTheta \in \setI$.

\subsection{Problem Statement}
We are interested in the problem of finding the parameters $\matTheta$ such that the forward kinematics model best explains a given set of 3D joint position observations. Let $\setJ \subset [J]$ denote the subset of joints for which the 3D positions $y_j \in \R^3, j \in \setJ$, are known. The IK problem can now be phrased as a (constrained) nonlinear least-squares problem that reads
\begin{align}
    &\min_{\matTheta \in \setI} \quad \sum_{j \in \setJ} \| y_j - x_j(\matTheta) \|^2 =: f(\matTheta)  \label{eq:ik}\\
    \Leftrightarrow  ~~&\min_{\matTheta \in \setI} \quad \sum_{j \in \setJ} \|\hat{y}_j - (\hat{t} + \left[ \prod_{i \in \alpha_j^1}  \hat{T}(a_i,\theta_i) \right] \hat{\zerovec}_3)\|^2 \,. \nonumber
\end{align}
Depending on the set $\setJ$, there may be multiple parameter vectors $\matTheta$ that all lead to the same configuration of joint positions $x_j(\matTheta)$ for all $j \in \setJ$. As such, for general IK problems of the form~\eqref{eq:ik}, the solution may not be unique since there can be multiple global optima.
Most commonly, such problems are solved based on local optimisation methods, e.g.~by modelling the hard constraints $\matTheta \in \setI$ as penalty in the objective function and then using a gradient descent procedure for locally optimising the objective function. A major downside of using such iterative approaches is that one requires a good initialisation for $\matTheta$, so that the optimisation does not result in an unwanted local optimum. We will tackle the problem of finding a good initialisation for Problem~\eqref{eq:ik} based on a convex relaxation, as we will describe next.

\section{Convex Relaxation for Inverse Kinematics}
In order to achieve a convex relaxation of the IK problem, we will first redefine the problem as a non-convex quadratically constrained quadratic programme (QCQP)~\cite{anstreicher2009semidefinite}. Subsequently, we will introduce our convex relaxation based on semidefinite programming.

\subsection{Inverse Kinematics as QCQP}\label{sec:ikasqcqp}
 Rather than phrasing the IK problem in terms of the parameter vector $\matTheta$, we will directly optimise for rotation matrices and the global translation.

\paragraph{Global and relative rotations:} 
Let $R_j \in \SO(3) \subset \R^{3 \times 3}$ denote the \emph{global} rotation of the $j$-th joint (i.e.~relative to the root), and let $R_j^{\local} \in \SO(3) \subset \R^{3 \times 3}$ denote the rotation of the $j$-th joint relative to its immediate parent, where we use the notation $\pi(j) \in [J]$ to indicate the parent of joint $j \in [J]$.  For all joints we have the relation 
\begin{align}
    R_j = R_{\pi(j)} R_j^{\local} \,,
\end{align}
where we define $R_{\pi(1)} = \matI_3$.
\paragraph{Forward kinematics:}
The position $p_j \in \R^3$ of the $j$-th (non-root) joint is defined recursively as
\begin{align}
    p_j = p_{\pi(j)} + R_{\pi(j)} v_j \,,
\end{align}
where $p_{\pi(j)} \in \R^3$ is the 3D position of its parent, and we define $p_{\pi(1)} := t$.

\paragraph{Joint angle constraints:}
For the $j$-th joint the rotation relative to its parent is constrained to be within the interval $\setI_j$, see Sec.~\ref{sec:fwkin}. In our reformulation we impose a similar constraint directly on the rotation matrix. 
    In order to do so, we express $R_j^{\local}$ in terms of a canonical rotation $R_j^{\canonical}$, where in our case we choose (w.l.o.g.) a rotation
    around the x-axis, so that we have the general structure
     \begin{align}\label{eq:rjcan}
        R_j^{\canonical} = \begin{bmatrix} 1& 0 & 0 \\ 0 & \cos(\theta_j) & -\sin(\theta_j) \\ 0 & \sin(\theta_j) & \cos(\theta_j)\end{bmatrix} =: \begin{bmatrix} 1& 0 & 0 \\ 0 & c_j & -s_j \\ 0 & s_j & c_j\end{bmatrix}\,.
    \end{align}
    As such, we can write $R_j^{\local} = S_j^T R_j^{\canonical} S_j$, for $S_j \in \SO(3)$ being a suitably chosen matrix that is determined a priori (i.e.~before optimisation). 
In order to impose the joint angle limits we enforce that $c_j$ and $s_j$ lie within the unit circle, i.e. we impose the convex constraint
\begin{align}
    c_j^2 + s_j^2 \leq 1\,.
\end{align}
In addition, for $\setI_j = [\theta_{j}^-, \theta_{j}^+]$,
we consider the line passing through $(\cos(\theta_{j}^-),\sin(\theta_{j}^-))$ and $(\cos(\theta_{j}^+),\sin(\theta_{j}^+))$, and enforce that the elements $c_j$ and $s_j$ of $R_j^{\canonical}$ are within a halfspace defined by this line. This results in a linear inequality constraint in $s_j$ and $c_j$. We use $R_j^{\canonical} \in \mathbb{L}_j$ to refer to both of these (convex) joint angle constraints.

\paragraph{Rotation constraints:}
The set of (proper) rotations $\SO(3)$ can be defined with quadratic constraints as 
\begin{align}
    \SO(3) = \{X \in \R^{3\times 3}~:~ X^TX = \matI_3, X_{:,1} {\times} X_{:,2} = X_{:,3}\} \,, \nonumber
\end{align}
where the cross-product is used to implement the right-hand rule in order to ensure that the determinant is $1$.

\paragraph{QCQP-IK:}
With the above elaborations we can now formulate the IK problem as the QCQP
\begin{align}
    &\min_{t, \{p_j,R_j,R_j^{\local},R_j^{\canonical}\}  } & &\sum_{j \in \setJ} \| y_j - p_j  \|^2  \label{eq:ik_reformulated}\\
    &\text{s.t.}  & & p_j = p_{\pi(j)} + R_{\pi(j)} v_j,~p_{\pi(1)} = t  \,,  \nonumber\\
    & & &R_j = R_{\pi(j)} R_j^{\local},~R_{\pi(1)} = \matI_3   \,,  \nonumber\\
    & & & R_j^{\local} = S_j^T R_j^{\canonical} S_j  \,, \nonumber\\
    & & & R_j^{\canonical} \in \mathbb{L}_j \,,  \nonumber\\
    & & & (R_j, R_j^{\local}, R_j^{\canonical}) \in \SO(3)^3   \,,  \nonumber
\end{align}
where the constraints are applied for all $j \in [J]$.
\subsection{Inverse Kinematics as SDP}\label{sec:sdpik}
Before we introduce our semidefinite programming relaxation of the IK problem, we briefly summarise the main idea of semidefinite programming relaxations for general QCQPs.

\subsubsection{Semidefinite Relaxations of Generic QCQPs}\label{sec:sdpqcqp}
A generic QCQP can be written in canonical form as
\begin{align}
    & \min_{x \in \R^n}  \quad &&x^T A_0 x \label{eq:qp}\\
    & \text{s.t.}  \quad  && x^T A_i x \leq b_i\,, \nonumber
\end{align}
where  $A_i \in \R^{n \times n}$ are given symmetric matrices (that are possibly indefinite). Note that by using a homogeneous coordinate representation, this form also allows for linear terms in the objective as well as for linear constraints. Commonly, such non-convex QCQPs are solved by means of lifting, where an additional lifted variable $Y$ of size $n{\times}n$ is introduced. Based on the property that for a given matrix $A$ it holds that $x^T A x = \trace(x^TAx) = \trace(Axx^T)$,  we can rewrite Problem~\eqref{eq:qp} as
\begin{align}
    & \min_{x \in \R^n, Y \in \R^{n {\times} n}}  \quad && \trace(A_0Y)\label{eq:qplifted}\\
    & \text{s.t.}  \quad  & &\trace(A_i Y) \leq b_i\,, ~ x x^T = Y\,. \nonumber
\end{align}
It is well-known that the constraint $xx^T = Y$ is equivalent to
\begin{align}
    \begin{bmatrix}
    1 & x^T \\ x & Y
    \end{bmatrix} \succeq 0
    ,~\operatorname{rank}(\begin{bmatrix}
    1 & x^T \\ x & Y
    \end{bmatrix}) = 1\,. \nonumber
\end{align}
Since the left part is a convex cone constraint, 
to obtain a convex relaxation the rank constraint (that accounts for the non-convexity) is dropped, which leads to the semidefinite programming problem
\begin{align}
    & \min_{x \in \R^n, Y \in \R^{n {\times} n}}  \quad && \trace(A_0Y)\label{eq:qpliftedcvx}\\
    & \text{s.t.}  \quad  & &\trace(A_i Y) \leq b_i \nonumber\\
    & & & \begin{bmatrix}
    1 & x^T \\ x & Y
    \end{bmatrix} \succeq 0 \,. \nonumber
\end{align}
\subsubsection{Semidefinite Relaxation for IK}
In order to obtain our semidefinite programming relaxation for the inverse kinematics problem, we systematically apply the elaborations in Sec.~\ref{sec:sdpqcqp} to Problem~\eqref{eq:ik_reformulated}. In the following we will elaborate on this.
\paragraph{Matrix multiplication constraints:} For three orthogonal matrices $X,Y,Z$ of size $n \times n$, the matrix constraint $X = YZ$ 
can equivalently be written as
\begin{align} \label{eq:multlifted}
    \begin{bmatrix} \matI_n & X\\ X^T & \matI_n \end{bmatrix} = \begin{bmatrix} Y \\ Z^T \end{bmatrix} \begin{bmatrix} Y \\ Z^T \end{bmatrix}^T\,.
\end{align}
Similarly as above, by dropping rank constraints, we obtain a convex relaxation of this constraint as
\begin{align}
    (X,Y,Z) \in \mathbb{M} := \{(X,Y,Z):\begin{bmatrix}   \matI_n & Y^T & Z \\ Y & \matI_n & X  \\ Z^T & X^T & \matI_n   \end{bmatrix} {\succeq} 0 \} \,. \nonumber
\end{align}
\paragraph{Rotation constraints:}  As indicated in Sec.~\ref{sec:ikasqcqp}, the $\SO(3)$ constraint can be represented with quadratic equality constraints. A semidefinite relaxation of the constraint $R \in \SO(3)$ is achieved by working with the vectorised $R$,~i.e.~$\VEC(R) \in \R^{9}$, and introducing a lifted variable $\mathbf{R}$ of size $9{\times}9$, as done in Sec.~\ref{sec:sdpqcqp}. The interested reader is referred to~\cite{Briales:un,Saunderson:2015fs}, where further details about the lifting and the constraints can be found. We use the notation $(R,\mathbf{R}) \in \mathbf{SO}(3)$, where $\mathbf{SO}(3)$ is a convex set, to indicate that $(R,\mathbf{R})$ is a pair of variables that satisfy the lifted rotation constraints.

\paragraph{Our convex relaxation:}
We now state the convex relaxation of the inverse kinematics problem, which reads
\begin{align}
    &\min_{\substack{t, \{p_j,R_j,R_j^{\local},R_j^{\canonical}\}\\
    \{\mathbf{R}_j,\mathbf{R}_j^{\local},\mathbf{R}_j^{\canonical}\}}  } & &\sum_{j \in \setJ} \| y_j - p_j  \|^2 \label{eq:cvxik}\\
    &\text{s.t.} & & p_j = p_{\pi(j)} + R_{\pi(j)} v_j,~p_{\pi(1)} = t  \,,  \nonumber\\
    & & &  (R_j, R_{\pi(j)}, R_j^{\local}) \in \mathbb{M}, ~R_{\pi(1)} = \matI_3  \,, \nonumber\\
    & & & R_j^{\local} = S_j^T R_j^{\canonical} S_j  \,,  \nonumber\\
    & & & R_j^{\canonical} \in \mathbb{L}_j \,,  \nonumber\\
    & & & (R_j, \mathbf{R}_j) \in \mathbf{SO}(3)   \,.  \nonumber\\
    & & & (R_j^{\local}, \mathbf{R}_j^{\local}) \in \mathbf{SO}(3)   \,.  \nonumber\\
    & & & (R_j^{\canonical}, \mathbf{R}_j^{\canonical}) \in \mathbf{SO}(3)   \,.  \nonumber
\end{align}

\paragraph{Practical considerations:} 
We solve our convex relaxation using the general purpose modelling tool Yalmip~\cite{lofberg2004yalmip} that interfaces the MOSEK solver~\cite{mosek}. In order to get a tighter convex relaxation it helps to introduce convex relaxations of redundant constraints,~e.g.~for orthogonality one would use convex relaxations for both the constraints $X^TX = \matI$ and $XX^T = \matI$, see~\cite{Briales:un}.
Once we have found a solution to Problem~\eqref{eq:cvxik}, we first project the obtained matrices onto the set $\SO(3)$ by means of Singular Value Decomposition, and subsequently extract the joint angles $\matTheta$ directly from the projected matrices.
The so-obtained joint angles $\matTheta$ are not necessarily an optimum of the original IK objective~\eqref{eq:ik}. Hence we use the found  $\matTheta$ as initialisation for a trust region method as implemented in Manopt~\cite{manopt}. We refer to our overall approach as \texttt{SDP-IK}.

\section{Experiments}
In this section we experimentally demonstrate the benefits of our proposed approach based on two real-world kinematic skeletons.
To this end, in Sec.~\ref{sec:local-opt} we first compare different local optimisation methods.
Subsequently, in Secs.~\ref{sec:noisefree-comp} and \ref{sec:noise-comp}, we compare the best-performing local optimisation method to our convex relaxation approach, where the effectiveness of the proposed \texttt{SDP-IK} will become apparent.
Since local IK methods are sensitive to initialisation, for every pose we uniformly sample multiple random initialisations (5 for the experiments in Sec.~\ref{sec:local-opt}, and 20 for the experiments in Sec.~\ref{sec:noisefree-comp} and~\ref{sec:noise-comp}) within the joint limit constraints, and then run local optimisation for each of the sampled initialisations. 
Note that we show the results obtained by \textit{all} random initialisations in the comparisons.
We measure the quality of the results based on the square-root of the normalised $f$ in \eqref{eq:ik}, which we denote as
\begin{equation}
\label{eq:ik-cost}
    \operatorname{cost}(\matTheta) = \sqrt{\frac{1}{|\setJ|} f(\matTheta)}\,.
\end{equation}
In our experiments, we use two real-world skeletons (\emph{hand} and human \emph{body}, see Fig.~\ref{fig:kinskel}) in 100 different natural poses sampled from captured motion data sequences~\cite{livecap2019,OccludedHands_ICCV2017}. 
The motion data is represented as a sequence of angle vectors $\matTheta$ that animate the hand and body kinematic skeletons with predefined bone vectors $\{v_j\}$.
We obtain the ``observed joint positions'' for each pose by computing the forward kinematics using the  angle vector $\matTheta$.
For the evaluation we consider two different cases:
\begin{enumerate}
    \item The observed joint positions are noise-free,~i.e.~there exist kinematic parameters that yield an exact fit with an objective value of $0$ (cf. Sec.~\ref{sec:noisefree-comp}).
    \item The observed joint positions have added noise,~i.e.~the existence of kinematic parameters for an exact fit is not guaranteed (cf. Sec.~\ref{sec:noise-comp}). This is the common case for practical applications.
\end{enumerate}

\subsection{Local Optimisation Methods}
\label{sec:local-opt}

\begin{figure}
\begin{tabular}{cc}
      \includegraphics[width=0.45\linewidth]{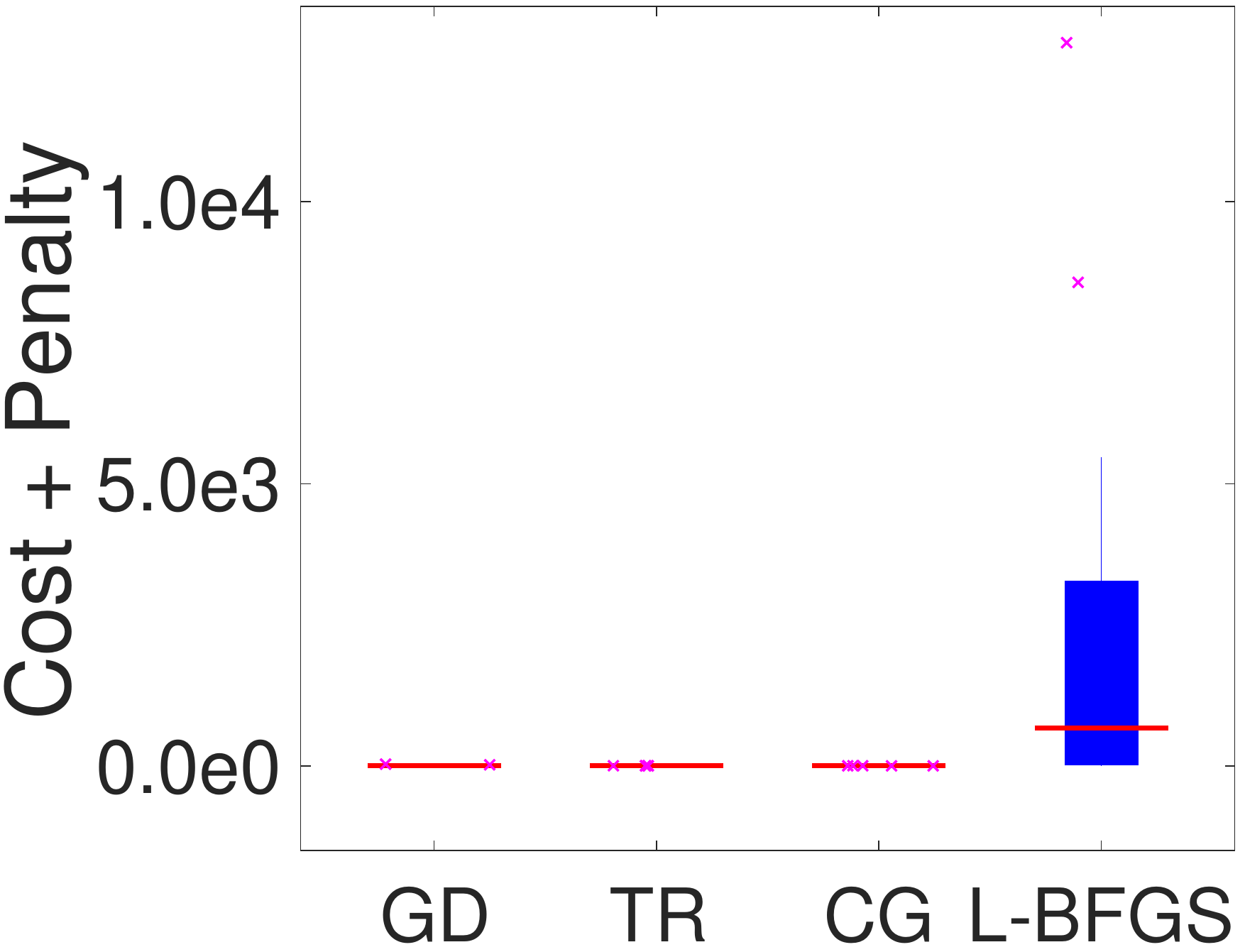} &   \includegraphics[width=0.45\linewidth]{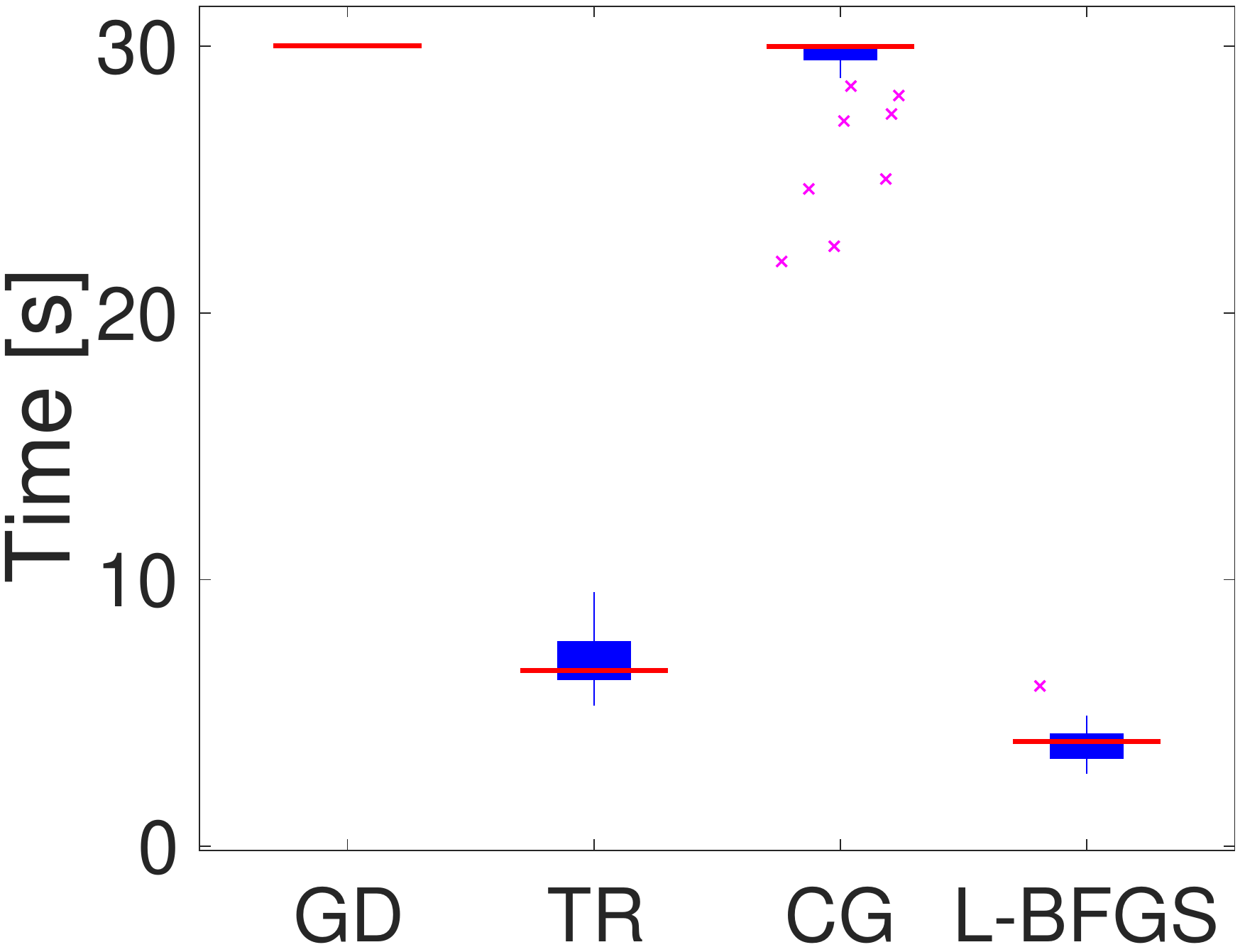} \\
    \end{tabular}
    \caption{We show the performance of different local optimisation methods: gradient descent (GD), trust region method (TR), conjugate gradient method (CG), and LBFGS.}
    \label{fig:ablation_localMethods} 
\end{figure}

To choose a representative local optimisation method as baseline, we compare four methods on the task of fitting the hand skeleton to observed joint locations when all  joints are observed (noise-free).
For tackling the IK problem with local optimisation methods it is common practice to convert it to an unconstrained optimisation problem, where the joint angle constraints are modelled as penalties. 
As such, we obtain the differentiable optimisation problem
\begin{equation}
    \min_{\matTheta \in \R^{3 + J}} \quad \sum_{j \in \setJ} \|y_j -  x_j(\matTheta)\|^2
    + \lambda \sum_{i = 1}^{3 + J} \text{dist}(\theta_i, \setI_i)^2 \, ,
\end{equation}
where $\text{dist}(\theta_i, \setI_i)$ measures the distance of $\theta_i$ from the interval of plausible angles $\setI_i$ (as done in \cite{GANeratedHands_CVPR2018}), and $\lambda$ is the hyperparameter that trades off joint limit violations with joint position errors. For our experiments we fix $\lambda=100$.

In Fig.~\ref{fig:ablation_localMethods} we show results of gradient descent (GD), the trust regions method (TR), conjugate gradient (CG), and limited memory Broyden-Fletcher-Goldfarb-Shanno (LBFGS) algorithms as implemented in the Manopt Matlab toolbox~\cite{manopt}. For all methods we used default parameters and allowed a maximum time budget of $30$s. Each method is tested with the same 8 poses with 5 random initialisations, so that a total of $40$ IK problems is solved. 
While the first three methods yield solutions with similar quality, LBFGS performs significantly worse.
Additionally, gradient descent and conjugate gradient are significantly slower compared to the other two methods.
Hence we decide to use the trust region method (TR) as representative local IK method for the remaining experiments due to its good trade-off between speed and accuracy.

\subsection{Fitting to Noise-Free Observations}
\label{sec:noisefree-comp}
\begin{figure}
    \begin{tabular}{cc}
      \includegraphics[width=0.45\linewidth]{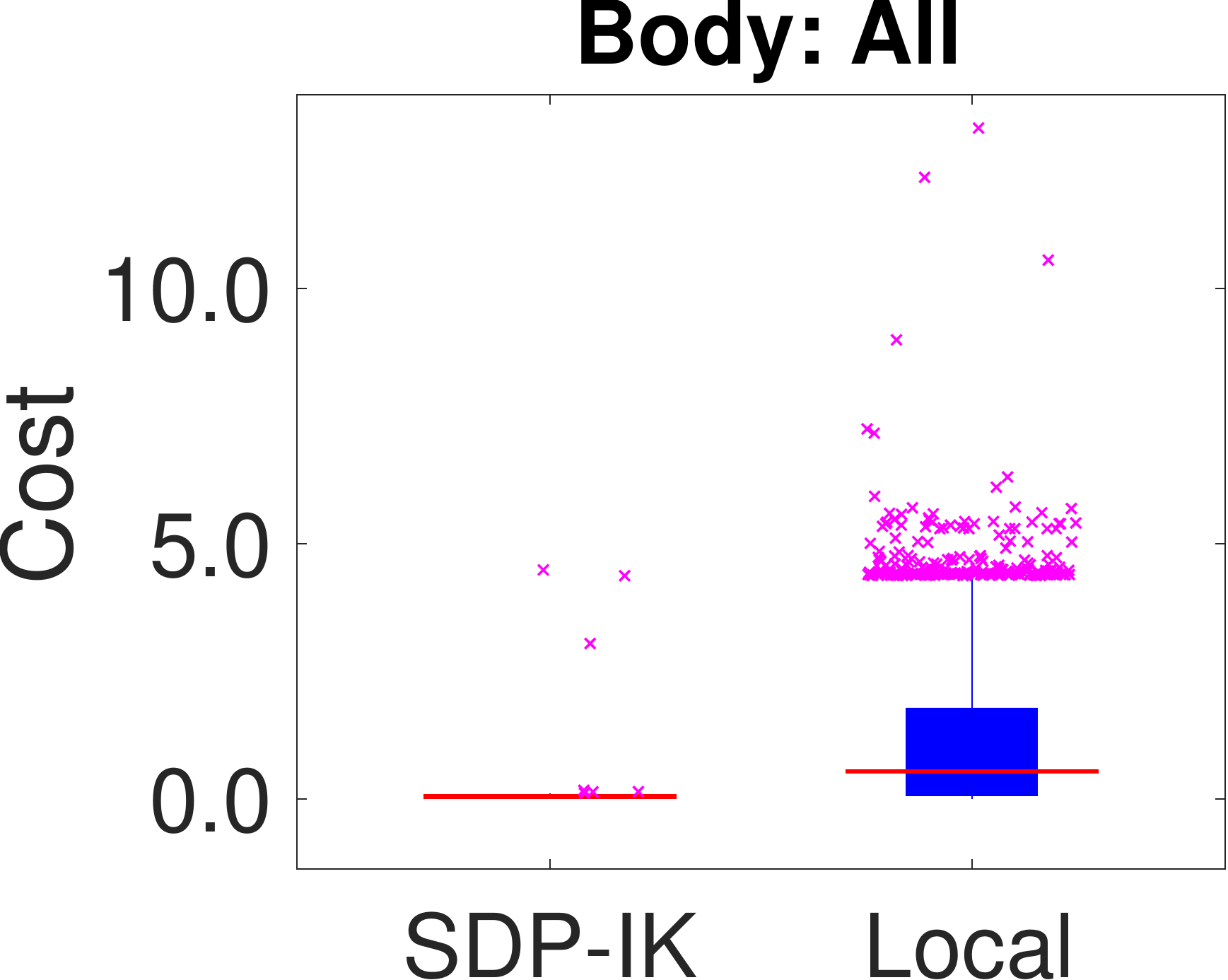} &   \includegraphics[width=0.45\linewidth]{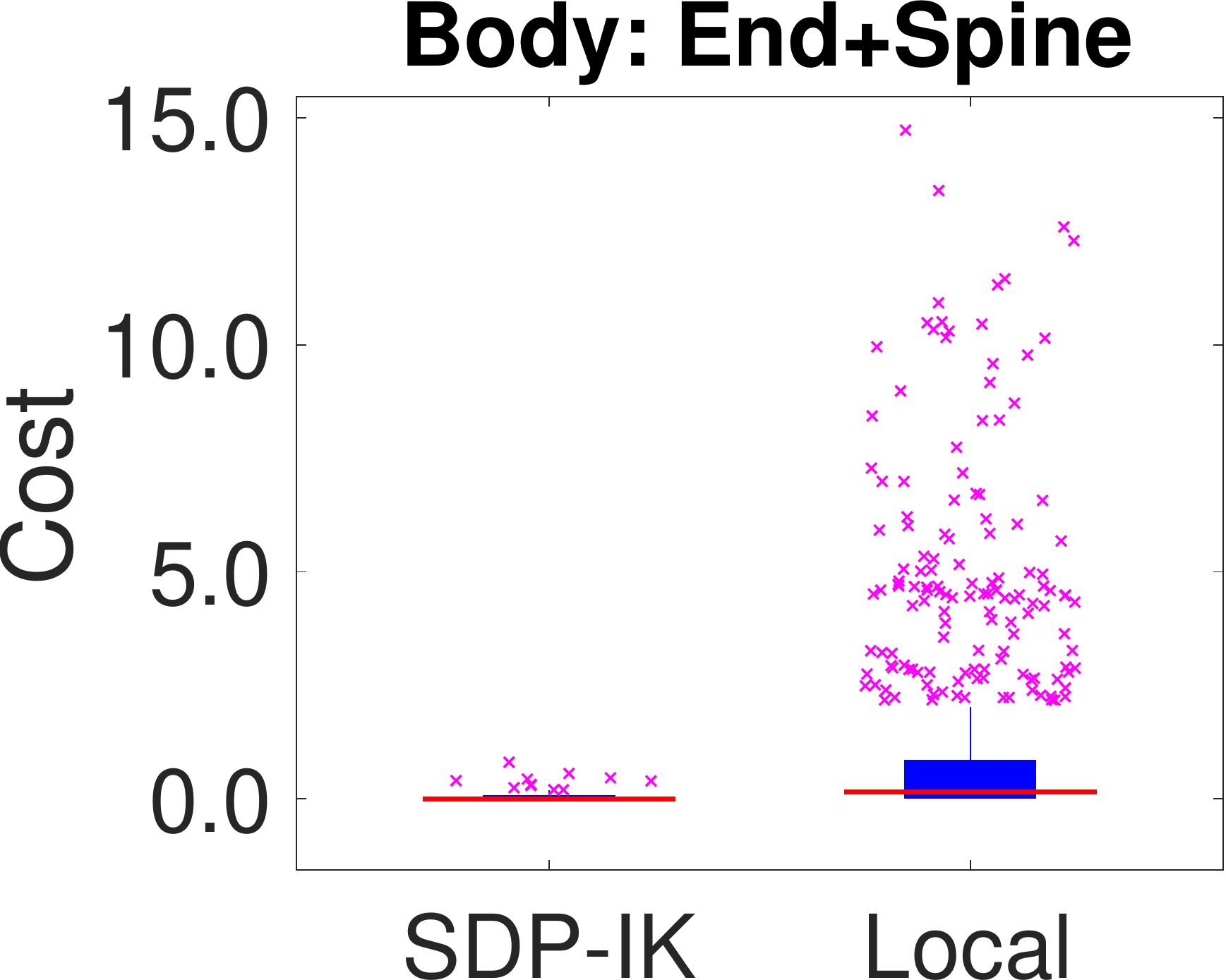} \\
     \includegraphics[width=0.45\linewidth]{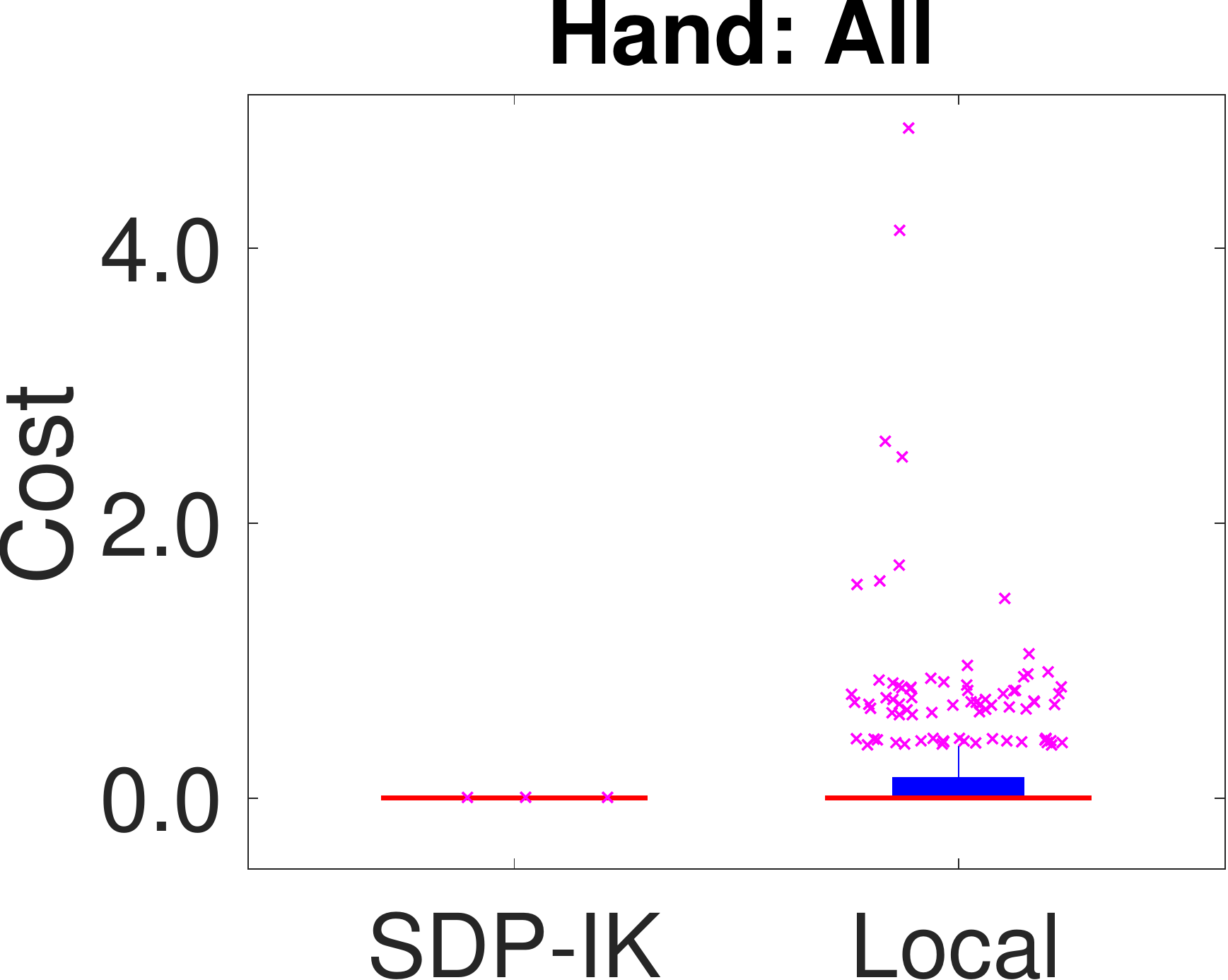} &   \includegraphics[width=0.45\linewidth]{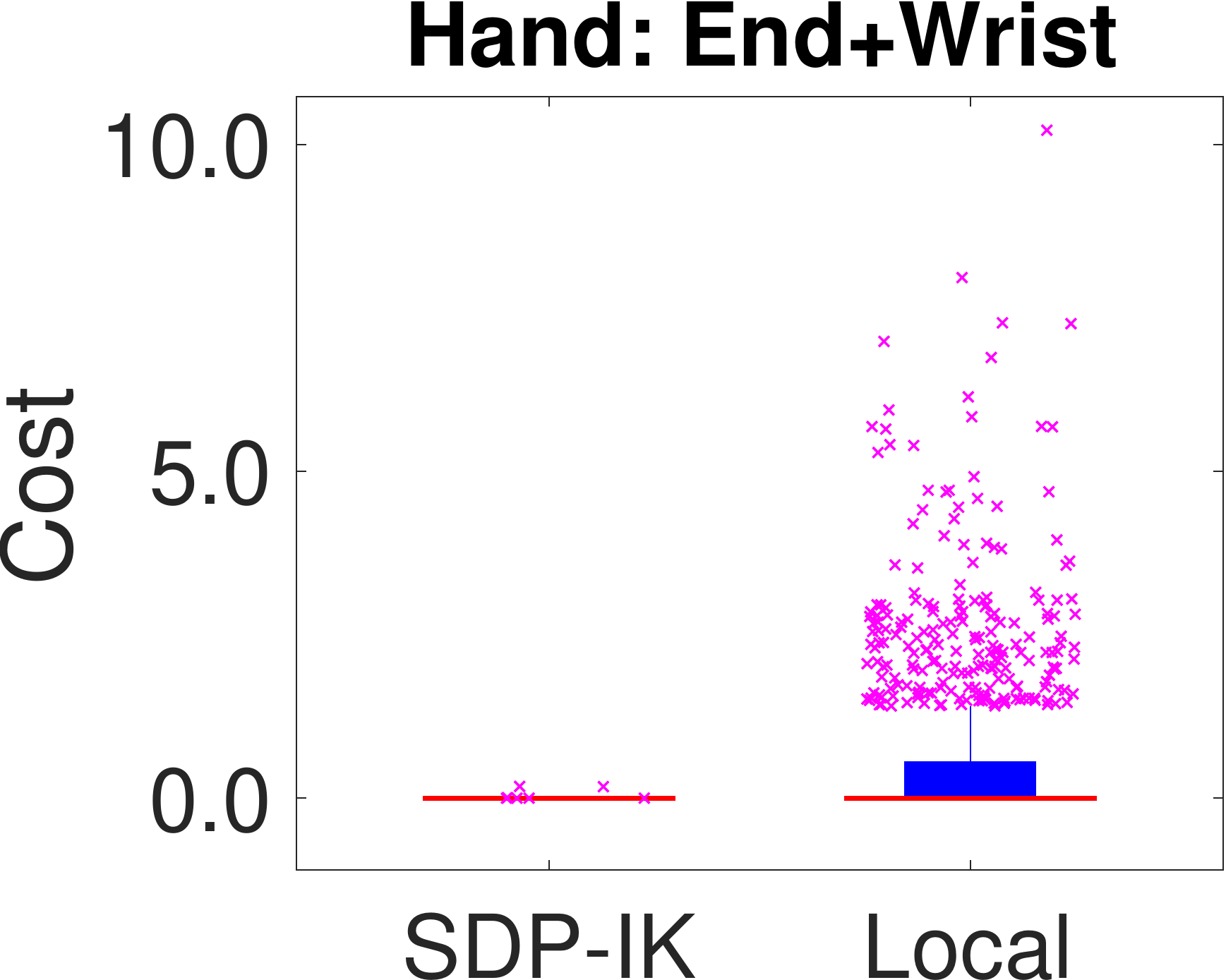} \\
    \end{tabular}
    \caption{Box plot for fitting to noise-free observed joint locations of 100 poses for different skeletons (top: \emph{body}, bottom: \emph{hand}). Our \texttt{SDP-IK} method yields a consistently lower error in terms of cost~\eqref{eq:ik-cost}, both in the case where all joints are given (top), as well as in the case when only the end-effectors are given (bottom).}
    \label{fig:clean_results_quant} 
\end{figure}

In Fig.~\ref{fig:clean_results_quant} we show quantitative results when solving the IK problem for noise-free observations.
Our \texttt{SDP-IK} method almost always achieves an exact fit, i.e.~the objective function value reaches 0, and thereby consistently outperforms the local optimisation approach. This occurs both when all joints are observed (\textbf{All}), and for the more ambiguous case where only the end-effectors and root are observed (\textbf{End+Spine} or \textbf{End+Wrist}), also cf.~Fig.~\ref{fig:kinskel}. Note that the local IK method suffers more from outliers due to bad initialisations, while our method can consistently find better minima.

\subsection{Fitting to Noisy Observations}
\label{sec:noise-comp}

In many situations a given inverse kinematics problem can potentially be infeasible,~i.e.~kinematic parameters that exactly explain given observed joint locations may not exist.
In practice, this is the case when the joint position observations are noisy.
For example, neural networks have been employed successfully for joint location prediction for hands or human bodies.
However, these networks mostly output a set of independent joint locations without any constraints to comply with the underlying kinematic structure.
Hence, some approaches subsequently fit a kinematic skeleton to the predictions to optimise for plausible joint angles, usually relying on local IK methods like gradient descent.
The aim then is to find kinematic parameters $\matTheta$ and hence a kinematically consistent and plausible set of joint positions that achieve a minimum distance to the observed locations, as~e.g.~done in~\cite{VNect_SIGGRAPH2017, GANeratedHands_CVPR2018}. 

In this experiment we mimic such a setting by fitting a kinematic skeleton to joint locations that exhibit \emph{noise}, so that an exact fit may not be possible.
To this end, we use the same poses as for the noise-free fitting in Sec.~\ref{sec:noisefree-comp}, and add noise to all observed joint locations.
For each joint, the 3D noise offset $n = r \cdot d \in \R^3$ is obtained by uniformly sampling a unit length direction $d \in \R^3$, and a magnitude $r \in \R$ from $[0, r_{\max}]$, where  $r_{\max}$ is $10$ mm for the hand skeleton, and $100$ mm for the body skeleton.

For infeasible cases, the minimum value of the IK cost function in~\eqref{eq:ik-cost} fluctuates randomly with the sampled noise. 
The resulting box plot would merely capture how much each noisy sample violates the kinematic constraints instead of the quality of the solution. 
Hence, instead of reporting the raw value of the IK cost, we provide  \emph{normalised cost} values.
To this end, for each pose we subtract the value of the smallest cost that is obtained by local optimisation among all the $20$ random initialisations.

\begin{figure}
    \begin{tabular}{cc}
      \includegraphics[width=0.45\linewidth]{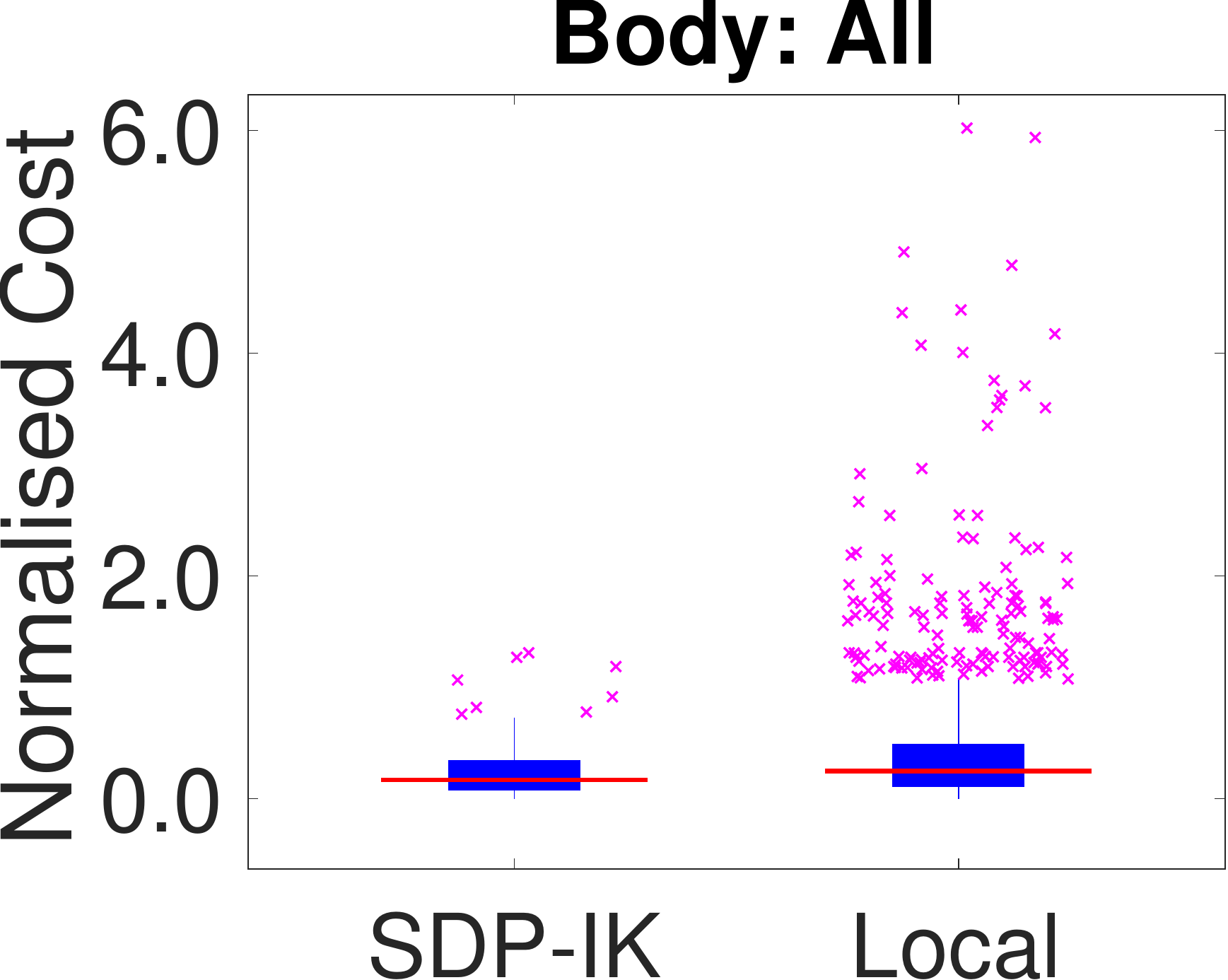} &   \includegraphics[width=0.45\linewidth]{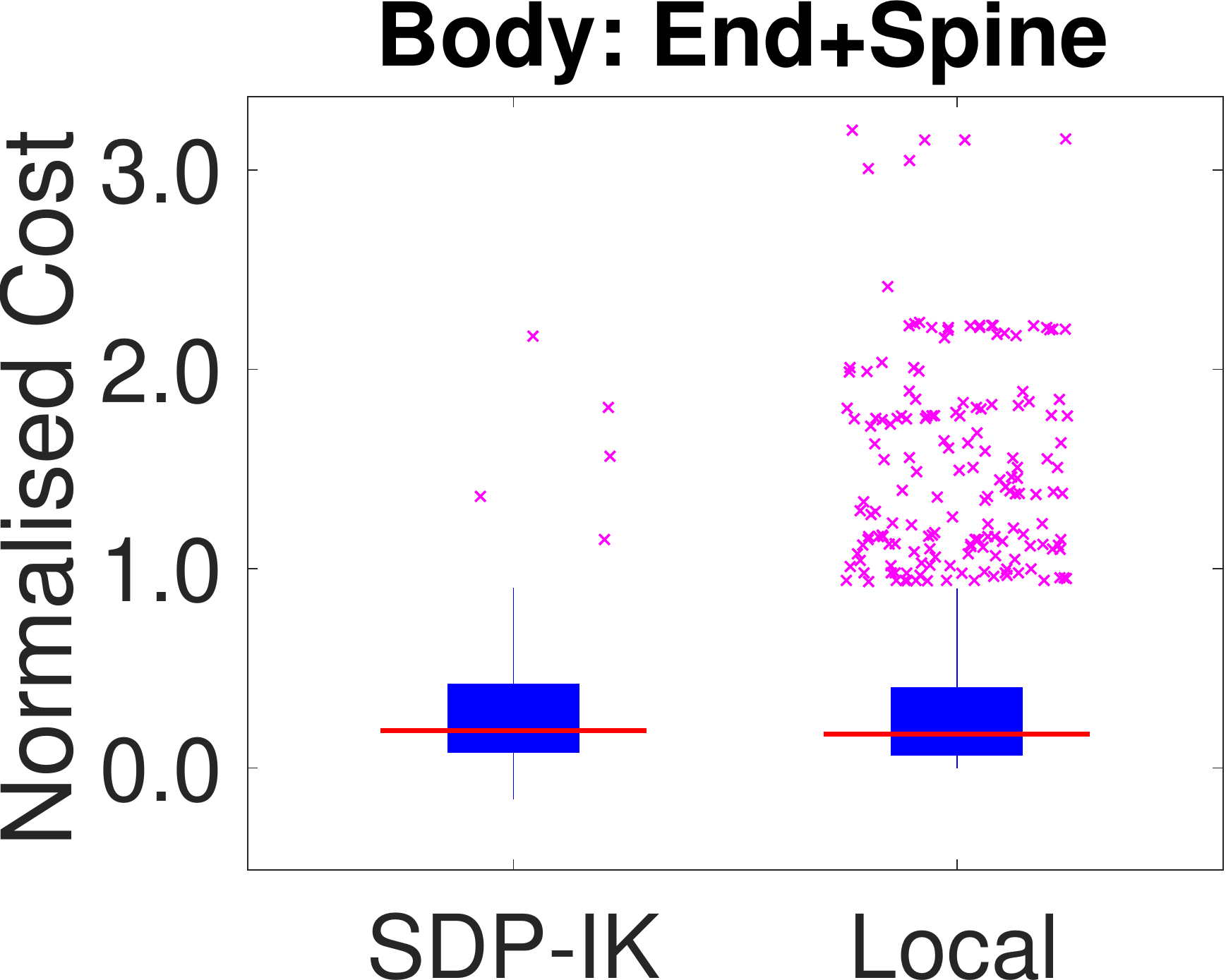} \\
     \includegraphics[width=0.45\linewidth]{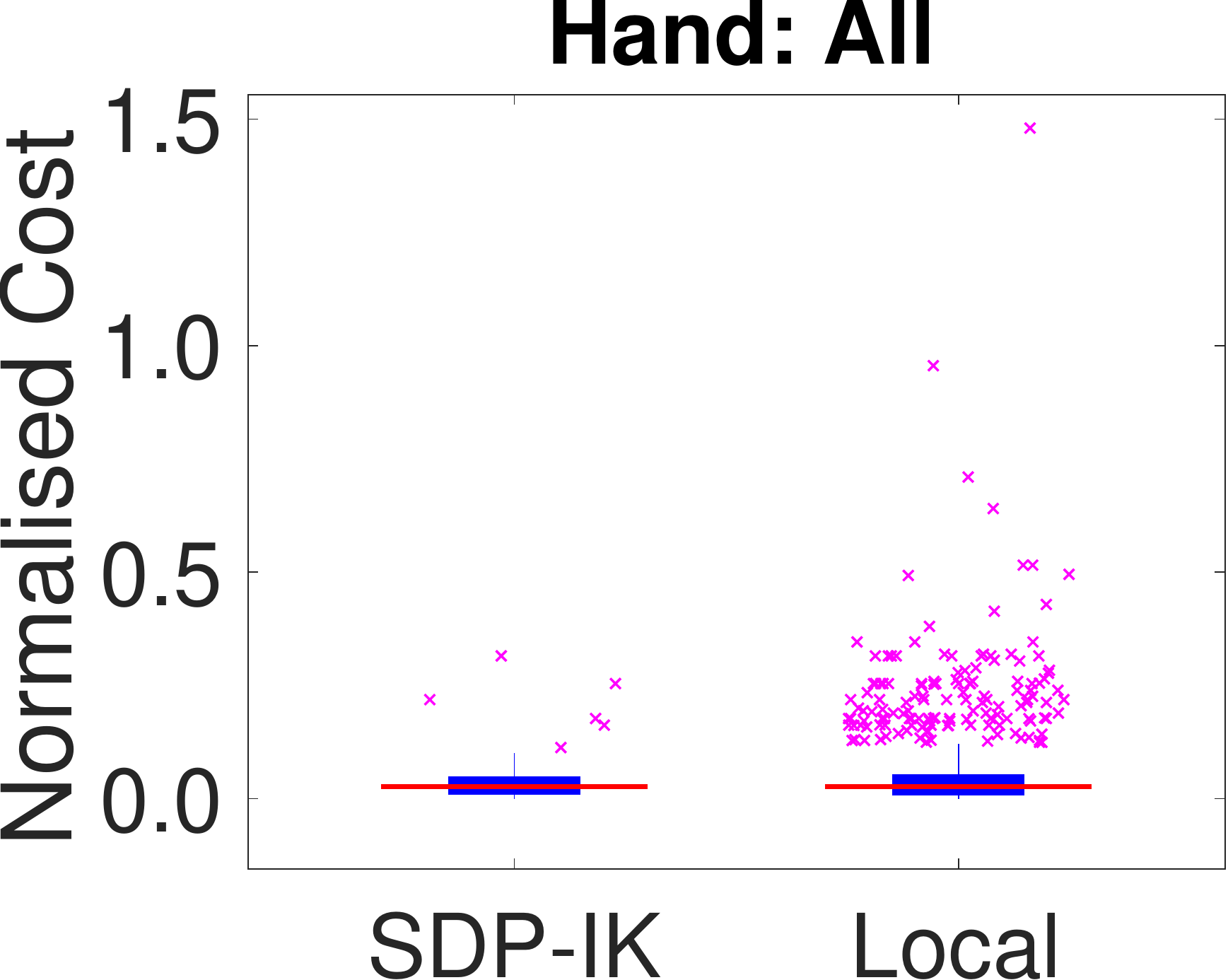} &   \includegraphics[width=0.45\linewidth]{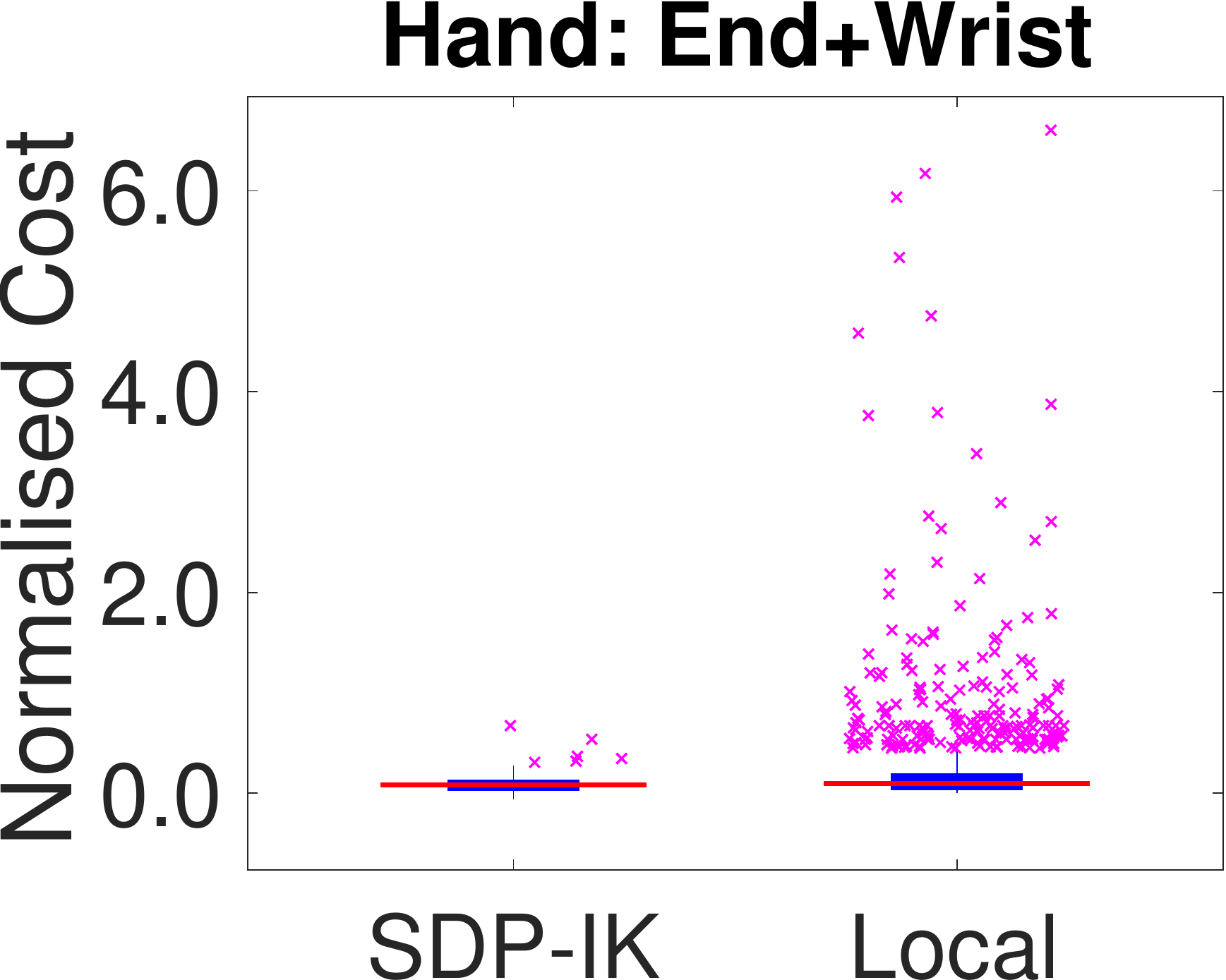} \\
    \end{tabular}
    \caption{Box plot for fitting to noisy observed joint locations of 100 poses for different skeletons (top: \emph{body}, bottom: \emph{hand}). Our \texttt{SDP-IK} method yields lower or similar error in terms of the normalised cost, both in the case where all joints are given (left), as well as in the case when only the end-effectors are given (right).}
    \label{fig:noisy_results_quant} 
\end{figure}

\begin{figure*}
    \includegraphics[width=\linewidth]{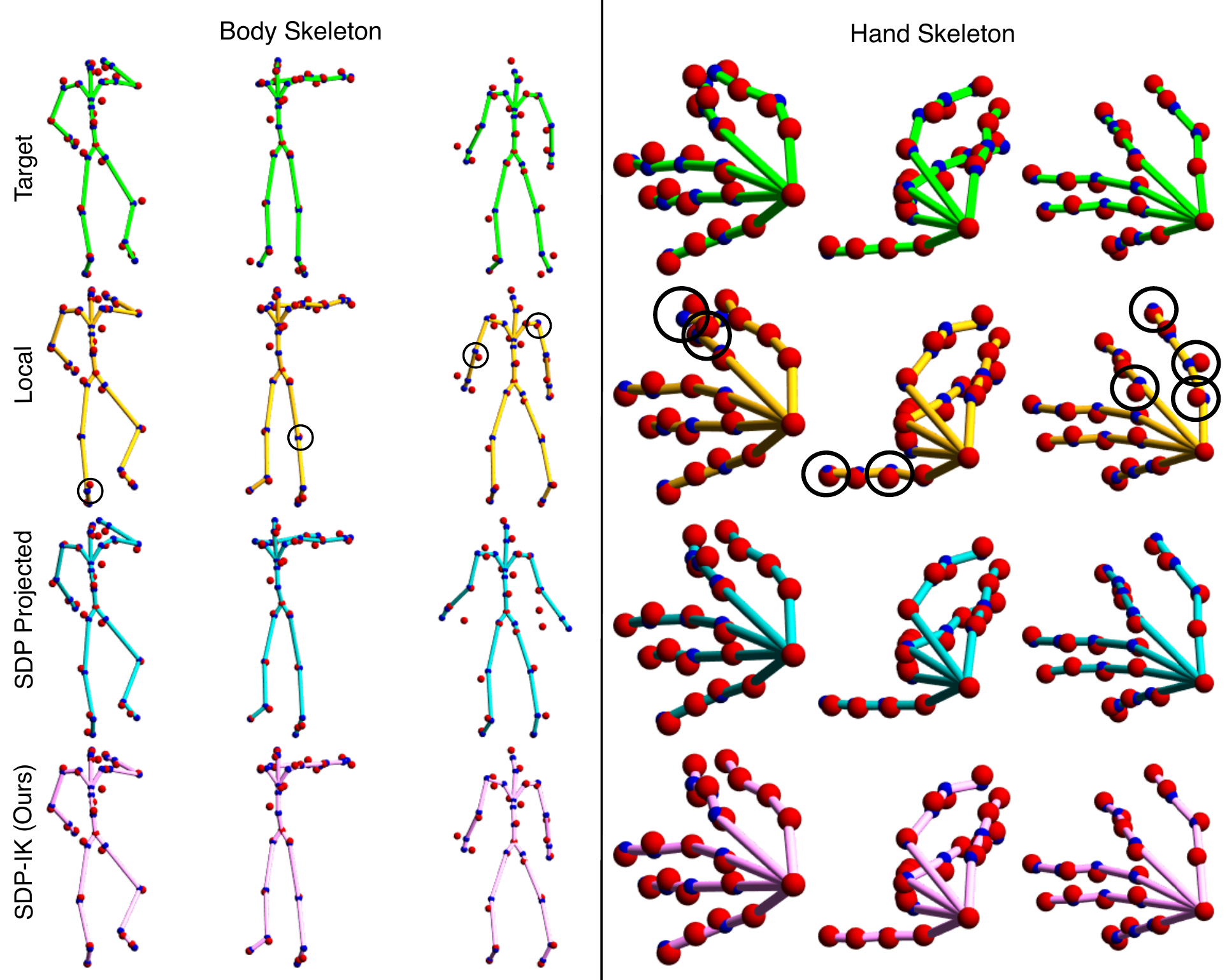}
    \caption{Qualitative results for fitting kinematic skeletons to known noisy points of human body and hand skeletons in different poses (red spheres: noisy points, blue spheres: fitted points). Top row: observed noisy joint positions. Second row: inverse kinematics solved via local optimisation using a random initialisation (erroneous joints are circled). Third row: solution of the semidefinite programming approach in \eqref{eq:cvxik} after projection. Bottom row: Our proposed \texttt{SDP-IK} method.}
    \label{fig:hand_motion_qual} 
\end{figure*}

In Fig.~\ref{fig:noisy_results_quant} we show that our method consistently achieves similar, or better performance than a local optimisation for the IK problem. Qualitative results for the local optimisation method, the projected solution of the semidefinite programming approach in~\eqref{eq:cvxik}, and our \texttt{SDP-IK} are shown in Fig.~\ref{fig:hand_motion_qual}.

\section{Discussion \& Limitations}
Currently, solving the semidefinite programming problem~\eqref{eq:cvxik}  takes about $17$s on average on an Intel i5 CPU with $8$GB of RAM, both for the hand as well as the human body skeleton. We emphasise that our current implementation is rather prototypical and is written so that it can work with generic kinematic skeletons. Moreover, we did not conduct any optimisations to improve the computational efficiency---we expect that runtime improvements by one order of magnitude or even more could be achieved using a problem-specific efficient implementation.

Currently, our model does not support translational DOF in non-root joints.
In the future, we plan to also look into this case, as translational DOF can account for small changes in the bone length, which in turn can deal with the variance of bone lengths in real-world data. 

\section{Conclusion}
We have presented a convex optimisation approach for the inverse kinematics problem based on a semidefinite programming relaxation. A major benefit of this approach is that we can find the global optimum of a (relaxation of the) IK problem, in contrast to local optimisation methods that heavily rely on good initialisations. Our experiments confirm this advantage and also demonstrate that our proposed \texttt{SDP-IK} approach is a useful method for tackling inverse kinematics problems as they appear in computer vision and computer graphic problems.

\subsection*{Acknowledgements}
This work was funded by the ERC Consolidator Grant 4DRepLy. We thank Marc Habermann for making the human body skeleton and motion sequences available to us.

{\small
\bibliographystyle{ieee}
\bibliography{references}

\begin{thebibliography}{10}\itemsep=-1pt

\bibitem{anstreicher2009semidefinite}
K.~M. Anstreicher.
\newblock Semidefinite programming versus the reformulation-linearization
  technique for nonconvex quadratically constrained quadratic programming.
\newblock {\em Journal of Global Optimization}, 43(2-3):471--484, 2009.

\bibitem{mosek}
M.~ApS.
\newblock {\em The MOSEK optimization toolbox for MATLAB manual. Version 9.0.},
  2019.

\bibitem{Aristidou2016}
A.~Aristidou, Y.~Chrysanthou, and J.~Lasenby.
\newblock Extending fabrik with model constraints.
\newblock {\em Computer Animation and Virtual Worlds}, 27:35--57, 02 2016.

\bibitem{Aristidou:2011:FABRIK}
A.~Aristidou and J.~Lasenby.
\newblock {FABRIK}: {A} fast, iterative solver for the inverse kinematics
  problem.
\newblock {\em Graph. Models}, 73(5):243--260, Sept. 2011.

\bibitem{Bandeira:2014wy}
A.~S. Bandeira, M.~Charikar, A.~Singer, and A.~Zhu.
\newblock {Multireference alignment using semidefinite programming.}
\newblock {\em ITCS}, 2014.

\bibitem{Abderrahim2013}
C.~{Bensalah}, J.~{Gonzalez-Quijano}, N.~{Hendrich}, and M.~{Abderrahim}.
\newblock Anthropomorphic robotics hand inverse kinematics using estimated svd
  in an extended sdls approach.
\newblock In {\em 2013 16th International Conference on Advanced Robotics
  (ICAR)}, pages 1--7, Nov 2013.

\bibitem{bernard:2018}
F.~Bernard, C.~Theobalt, and M.~Moeller.
\newblock {DS*: Tighter Lifting-Free Convex Relaxations for Quadratic Matching
  Problems}.
\newblock In {\em CVPR}, 2018.

\bibitem{manopt}
N.~Boumal, B.~Mishra, P.-A. Absil, and R.~Sepulchre.
\newblock {M}anopt, a {M}atlab toolbox for optimization on manifolds.
\newblock {\em Journal of Machine Learning Research}, 15:1455--1459, 2014.

\bibitem{Briales:un}
J.~Briales and J.~Gonzalez-Jimenez.
\newblock {Convex Global 3D Registration with Lagrangian Duality.}
\newblock {\em CVPR}, 2017.

\bibitem{BussIK}
S.~Buss.
\newblock Introduction to inverse kinematics with jacobian transpose,
  pseudoinverse and damped least squares methods.
\newblock {\em IEEE Transactions in Robotics and Automation}, 17, 05 2004.

\bibitem{Bocsi:STRUCTIK:2011}
B.~{Bócsi}, D.~{Nguyen-Tuong}, L.~{Csató}, B.~{Schölkopf}, and J.~{Peters}.
\newblock Learning inverse kinematics with structured prediction.
\newblock In {\em 2011 IEEE/RSJ International Conference on Intelligent Robots
  and Systems}, pages 698--703, Sep. 2011.

\bibitem{Carlone:2018ji}
L.~Carlone and G.~C. Calafiore.
\newblock {Convex Relaxations for Pose Graph Optimization With Outliers.}
\newblock {\em IEEE Robotics and Automation Letters}, 2018.

\bibitem{ChenGuibasHuang14_NearOptimalJointObjectMatchingViaConvexRelaxation}
Y.~Chen, L.~J. Guibas, and Q.-X. Huang.
\newblock {Near-Optimal Joint Object Matching via Convex Relaxation}.
\newblock In {\em International Conference on Machine Learning (ICML)}, 2014.

\bibitem{Torras2012}
A.~{Colomé} and C.~{Torras}.
\newblock Redundant inverse kinematics: Experimental comparative review and two
  enhancements.
\newblock In {\em 2012 IEEE/RSJ International Conference on Intelligent Robots
  and Systems}, pages 5333--5340, Oct 2012.

\bibitem{dai2017global}
H.~Dai, G.~Izatt, and R.~Tedrake.
\newblock Global inverse kinematics via mixed-integer convex optimization.
\newblock In {\em International Symposium on Robotics Research, Puerto Varas,
  Chile}, pages 1--16, 2017.

\bibitem{dai2018synthesis}
H.~Dai, A.~Majumdar, and R.~Tedrake.
\newblock Synthesis and optimization of force closure grasps via sequential
  semidefinite programming.
\newblock In {\em Robotics Research}, pages 285--305. Springer, 2018.

\bibitem{Eriksson:2018cq}
A.~P. Eriksson, C.~Olsson, F.~Kahl, and T.-J. Chin.
\newblock {Rotation Averaging and Strong Duality.}
\newblock {\em CVPR}, 2018.

\bibitem{livecap2019}
M.~Habermann, W.~Xu, M.~Zollhoefer, G.~Pons-Moll, and C.~Theobalt.
\newblock Livecap: Real-time human performance capture from monocular video,
  2019.

\bibitem{Hecker:2008:RMR:1399504.1360626}
C.~Hecker, B.~Raabe, R.~W. Enslow, J.~DeWeese, J.~Maynard, and K.~van Prooijen.
\newblock Real-time motion retargeting to highly varied user-created
  morphologies.
\newblock In {\em ACM SIGGRAPH 2008 Papers}, SIGGRAPH '08, pages 27:1--27:11,
  New York, NY, USA, 2008. ACM.

\bibitem{Kenwright2012}
B.~{Kenwright}.
\newblock Real-time character inverse kinematics using the gauss-seidel
  iterative approximation method.
\newblock In {\em 2012 4th International Conference on Creative Content
  Technologies (CONTENT 2012)}, pages 63--68, 2012.

\bibitem{kezurer2015}
I.~Kezurer, S.~Z. Kovalsky, R.~Basri, and Y.~Lipman.
\newblock {Tight Relaxation of Quadratic Matching.}
\newblock {\em Comput. Graph. Forum}, 2015.

\bibitem{Khoo:2016hr}
Y.~Khoo and A.~Kapoor.
\newblock {Non-Iterative Rigid 2D/3D Point-Set Registration Using Semidefinite
  Programming.}
\newblock {\em IEEE Trans. Image Processing}, 2016.

\bibitem{Chin97}
{Kwan Wu Chin}, B.~R. {von Konsky}, and A.~{Marriott}.
\newblock Closed-form and generalized inverse kinematics solutions for the
  analysis of human motion.
\newblock In {\em Proceedings of the 19th Annual International Conference of
  the IEEE Engineering in Medicine and Biology Society. 'Magnificent Milestones
  and Emerging Opportunities in Medical Engineering' (Cat. No.97CH36136)},
  volume~5, pages 1911--1914 vol.5, Oct 1997.

\bibitem{lofberg2004yalmip}
J.~Lofberg.
\newblock Yalmip: A toolbox for modeling and optimization in matlab.
\newblock In {\em IEEE international conference on robotics and automation},
  pages 284--289, 2004.

\bibitem{Maron:2016vv}
H.~Maron, N.~Dym, I.~Kezurer, S.~Kovalsky, and Y.~Lipman.
\newblock {Point registration via efficient convex relaxation}.
\newblock {\em ACM Transactions on Graphics (TOG)}, 35(4):73, 2016.

\bibitem{VNect_SIGGRAPH2017}
D.~Mehta, S.~Sridhar, O.~Sotnychenko, H.~Rhodin, M.~Shafiei, H.-P. Seidel,
  W.~Xu, D.~Casas, and C.~Theobalt.
\newblock Vnect: Real-time 3d human pose estimation with a single rgb camera.
\newblock volume~36, 2017.

\bibitem{Merrick:2004}
D.~Merrick and T.~Dwyer.
\newblock Skeletal animation for the exploration of graphs.
\newblock In {\em Proceedings of the 2004 Australasian Symposium on Information
  Visualisation - Volume 35}, APVis '04, pages 61--70, Darlinghurst, Australia,
  Australia, 2004. Australian Computer Society, Inc.

\bibitem{GANeratedHands_CVPR2018}
F.~Mueller, F.~Bernard, O.~Sotnychenko, D.~Mehta, S.~Sridhar, D.~Casas, and
  C.~Theobalt.
\newblock {GANerated Hands for Real-Time 3D Hand Tracking from Monocular RGB}.
\newblock In {\em Proceedings of Computer Vision and Pattern Recognition
  ({CVPR})}, June 2018.

\bibitem{OccludedHands_ICCV2017}
F.~Mueller, D.~Mehta, O.~Sotnychenko, S.~Sridhar, D.~Casas, and C.~Theobalt.
\newblock Real-time hand tracking under occlusion from an egocentric rgb-d
  sensor.
\newblock In {\em Proceedings of International Conference on Computer Vision
  ({ICCV})}, 2017.

\bibitem{Rosen:2015dv}
D.~M. Rosen, C.~DuHadway, and J.~J. Leonard.
\newblock {A convex relaxation for approximate global optimization in
  simultaneous localization and mapping.}
\newblock {\em ICRA}, 2015.

\bibitem{Saunderson:2015fs}
J.~Saunderson, P.~A. Parrilo, and A.~S. Willsky.
\newblock {Semidefinite Descriptions of the Convex Hull of Rotation Matrices.}
\newblock {\em SIAM Journal on Optimization}, 2015.

\bibitem{Schellewald:2005up}
C.~Schellewald and C.~Schn{\"o}rr.
\newblock {Probabilistic subgraph matching based on convex relaxation}.
\newblock In {\em EMMCVPR}, 2005.

\bibitem{ellipsoidtracker_3dv2014}
S.~Sridhar, H.~Rhodin, H.-P. Seidel, A.~Oulasvirta, and C.~Theobalt.
\newblock Real-time hand tracking using a sum of anisotropic gaussians model.
\newblock In {\em Proceedings of the International Conference on 3D Vision
  ({3DV})}, Dec. 2014.

\bibitem{starke2016efficient}
S.~Starke, N.~Hendrich, S.~Magg, and J.~Zhang.
\newblock An efficient hybridization of genetic algorithms and particle swarm
  optimization for inverse kinematics.
\newblock In {\em 2016 IEEE International Conference on Robotics and
  Biomimetics (ROBIO)}, pages 1782--1789, 2016.

\bibitem{starke2017memetic}
S.~Starke, N.~Hendrich, and J.~Zhang.
\newblock A memetic evolutionary algorithm for real-time articulated kinematic
  motion.
\newblock In {\em IEEE Congress on Evolutionary Computation (CEC)}, pages
  2473--2479, 2017.

\bibitem{Tkach:2016}
A.~Tkach, M.~Pauly, and A.~Tagliasacchi.
\newblock Sphere-meshes for real-time hand modeling and tracking.
\newblock {\em ACM Trans. Graph.}, 35(6):222:1--222:11, Nov. 2016.

\bibitem{Wang:2013tq}
L.~Wang and A.~Singer.
\newblock {Exact and stable recovery of rotations for robust synchronization}.
\newblock {\em Information and Inference}, 2013.

\bibitem{Wang1991}
L.~.~T. {Wang} and C.~C. {Chen}.
\newblock A combined optimization method for solving the inverse kinematics
  problems of mechanical manipulators.
\newblock {\em IEEE Transactions on Robotics and Automation}, 7(4):489--499,
  Aug 1991.

\bibitem{Wang:2013vq}
P.~Wang, C.~Shen, and A.~van~den Hengel.
\newblock {A Fast Semidefinite Approach to Solving Binary Quadratic Problems}.
\newblock {\em CVPR}, 2013.

\bibitem{Zhao:1998wc}
Q.~Zhao, S.~E. Karisch, F.~Rendl, and H.~Wolkowicz.
\newblock {Semidefinite programming relaxations for the quadratic assignment
  problem}.
\newblock {\em Journal of Combinatorial Optimization}, 2(1):71--109, 1998.

\end{thebibliography}
}

\end{document}